\pdfoutput=1
\documentclass{article}
\PassOptionsToPackage{numbers,sort&compress}{natbib}
\newif\ifpreprint
\ifdefined\submissionmode\preprintfalse\else\preprinttrue\fi
\ifpreprint
  \usepackage{dualverse}
\else
  \usepackage[main]{neurips_2025}
\fi
\usepackage[utf8]{inputenc}
\usepackage[T1]{fontenc}
\ifpreprint\usepackage{lmodern}\fi
\usepackage{hyperref}
\usepackage{url}
\usepackage{booktabs}
\usepackage{longtable}
\usepackage{amsfonts,amsmath,amssymb,bm}
\usepackage{nicefrac}
\usepackage{microtype}
\usepackage{xcolor}
\usepackage{graphicx}
\usepackage{subcaption}
\usepackage{multirow}
\usepackage{float}
\usepackage{needspace}
\usepackage{placeins}
\usepackage{enumitem}
\usepackage{tikz}
\usetikzlibrary{arrows.meta,positioning,calc,fit,backgrounds}
\usepackage[capitalise,noabbrev]{cleveref}

\ifpreprint
  \newcommand{\ChorusHeadRow}{\DualverseHeadRow}
  \newcommand{\ChorusHead}[1]{\DualverseHead{#1}}
  \AtBeginEnvironment{table}{\renewcommand{\arraystretch}{1.0}}
\else
  \newcommand{\ChorusHeadRow}{}
  \newcommand{\ChorusHead}[1]{\textbf{#1}}
\fi

\newcommand{\method}{Chorus}

\newcommand{\bpb}{BPB}

\newcommand{\Wikidataengram}{17.38 $\pm$ 1.20}

\newcommand{\Wikidatatngram}{18.57 $\pm$ 1.28}

\newcommand{\Scorecodem}{0.1240 $\pm$ 0.0016}
\newcommand{\Wikidatacodem}{25.84 $\pm$ 2.74}
\newcommand{\CtlSharedBigramAdamw}{0.9645 $\pm$ 0.0004}
\newcommand{\CtlEngramEmber}{0.9575 $\pm$ 0.0004}

\newcommand{\CtlFEEmber}{0.9860 $\pm$ 0.0017}

\newcommand{\CtlOursAdamw}{0.9493 $\pm$ 0.0012}

\newcommand{\CtlSharedBigram}{0.9622 $\pm$ 0.0006}

\newcommand{\CtlEngramNativeB}{0.9616 $\pm$ 0.0006}
\newcommand{\CtlFENative}{0.9565 $\pm$ 0.0005}
\newcommand{\CtlOursEmber}{0.9473 $\pm$ 0.0005}

\newcommand{\norm}{\operatorname{Norm}}
\newcommand{\silu}{\operatorname{SiLU}}
\newcommand{\sg}{\operatorname{sg}}
\newcommand{\R}{\mathbb{R}}
\DeclareMathOperator{\conv}{Conv}

\definecolor{oursblue}{HTML}{2A78D6}
\definecolor{engramorange}{HTML}{EB6834}
\definecolor{feaqua}{HTML}{1BAF7A}
\definecolor{tnyellow}{HTML}{EDA100}
\definecolor{ink}{HTML}{1F2328}
\definecolor{muted}{HTML}{5B616A}
\definecolor{tablefill}{HTML}{E4EEFB}
\definecolor{sitefill}{HTML}{F3F4F6}

\title{\method{}: Shared N-gram Embeddings\texorpdfstring{\\}{ }with Layer-wise Adaptation}

\ifpreprint
  \SetPaperDate{18-Nov-2026}
\SetCorrespondence{info@dualverse.ai}
\SetPaperLogo{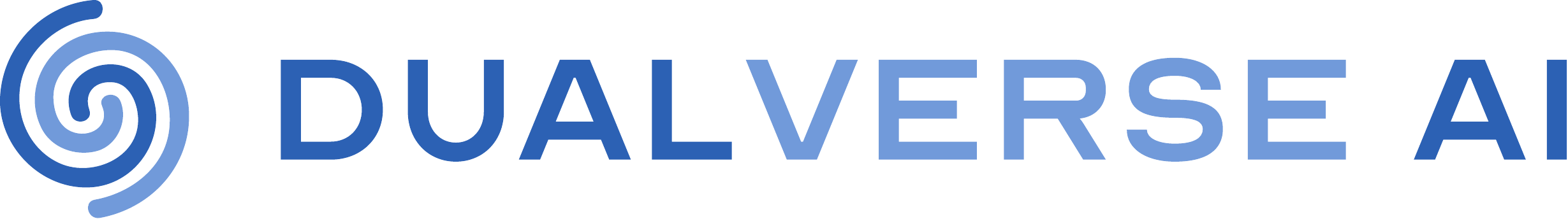}
\SetPDFAuthor{Stephen Chung, Wenyu Du}
\SetHeaderTitle{Chorus: Shared N-gram Embeddings with Layer-wise Adaptation}
\SetDualverseAuthors{%
  Stephen Chung\\DualverseAI%
  \and Wenyu Du\\DualverseAI%
}{%
  {\sffamily\bfseries\color{DualverseA}%
    Stephen Chung \quad\quad Wenyu Du%
  }\\[1em]
  \parbox{0.9\linewidth}{\centering
    \normalfont\sffamily\bfseries\small\color{DualverseB}%
    DualverseAI%
  }%
}

\else
  \author{Anonymous Author(s)}
\fi

\begin{document}
\ifpreprint\raggedbottom\fi
\typeout{FORMAT-LINEWIDTH=\the\linewidth}
\ifpreprint\else\maketitle\fi

\begin{abstract}
Language models can gain capacity at little computational cost from n-gram embedding tables: large tables of learned vectors for short sequences of consecutive tokens, which are looked up and added to the model's input or to its intermediate layers.
Recent designs commonly use a separate n-gram embedding table for each layer, which makes the tables costly in memory and prevents layers from sharing what they learn.
However, a single shared table gives every layer the same embedding, leaves no room for layer-wise preferences, and performs poorly in practice.
We propose \method{}, in which one set of n-gram embedding tables serves the whole network and every layer reads it in its own way.
Each layer maps the refined embedding through its own projection layer, after rescaling its components by factors computed from its current hidden state.
\method{} matches the quality of existing n-gram methods with 87\% smaller tables: with 1.2B table parameters it predicts text better than FactorEngram and DeepSeek's Engram do with 9.6B.
It outperforms Engram, TN-gram and FactorEngram in every setting where we compared them, across table sizes from 20M to 9.6B parameters.
We also show that the recently proposed memory-efficient optimizer Ember works well with \method{}, saving 99.2\% of table-optimizer memory compared with AdamW without degrading training performance.
These results enable further scaling of n-gram embedding tables along the table-size axis by reducing memory use and allowing shared knowledge to be reused across layers.
\ifpreprint
\par\smallskip
\noindent\textbf{Code:} \url{https://github.com/dualverse-ai/chorus}
\fi
\end{abstract}
\ifpreprint
\maketitle
\begingroup
\renewcommand{\thefootnote}{}
\footnotetext[1]{This work extends a human-verified finding made by an rStation agent (\mbox{\url{https://rstation.ai/}}) as part of our \href{https://dualverse.ai/auto_research}{Autonomous Research Program} (\mbox{\url{https://dualverse.ai/auto_research}}).}
\endgroup
\vspace{1.8ex}
\fi

\section{Introduction}
\label{sec:intro}

In a standard language model, model size and computation grow together because almost every weight takes part in processing every token \citep{vaswani2017,kaplan2020,hoffmann2022}.
N-gram embedding tables separate the two by adding large tables of learned vectors for short sequences of consecutive tokens.
At each position, the model looks up vectors for the short sequences ending at the current token, such as the two-token bigram and three-token trigram.
Because far too many such sequences exist to give each its own row, a fixed hash function maps each sequence to a row of the table \citep{weinberger2009,svenstrup2017}.
Each position reads only a few rows however large the tables are, so their size is decoupled from the computation each token costs, and the tables can hold billions of parameters at little computational cost.

Earlier models placed n-gram embedding tables only at the input \citep{huang2025oe,yu2025scone}, whereas recent designs place them inside the network so that looked-up vectors interact with intermediate representations.
DeepSeek's Engram inserts hashed n-gram tables into selected layers and gates the retrieved vectors with the layer's hidden state \citep{engram}.
Follow-up designs change how each table is structured: TN-gram shares tensor factors across n-gram orders \citep{tngram}, FactorEngram stores sparse coefficients over a learned dictionary \citep{factorengram}, and X-gram allocates rows by token frequency \citep{xgram}.
Despite these differences, these methods commonly use a separate n-gram embedding table for each layer (\cref{fig:method}a).
Separate tables are costly in memory: reading n-grams at more layers either multiplies table size or leaves fewer rows per table and more hash collisions.
They also prevent layers from sharing what they learn about the same token sequences.

\begin{figure}[t]
\centering
\ifdualversehtml
\includegraphics[width=\textwidth]{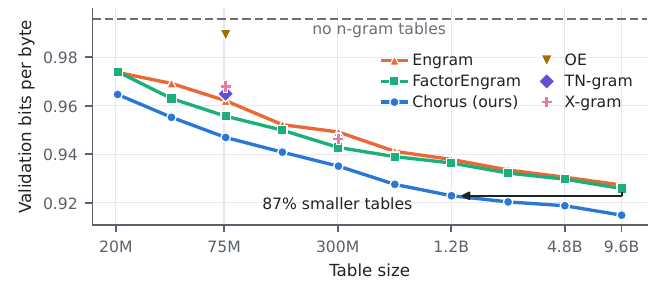}
\else
\includegraphics[width=0.8\linewidth]{figures/fig_scaling.pdf}
\fi
\caption{\textbf{Scaling curves with respect to table size.} ClimbMix, nine-layer model (20.7M backbone), ten table sizes. Validation bits per byte (lower is better) against table size. Curves show \method{}, Engram and FactorEngram; isolated points show OE, TN-gram and two X-gram configurations. One seed per point.}
\label{fig:scaling}
\end{figure}

Sharing a single table across layers is the natural alternative, and simple forms of it exist.
Compact training recipes add token-indexed vectors to the attention values of several layers \citep{jordan2024moddednanogpt,karpathy2025nanochat}.
A single hashed bigram table has likewise been shared across layers \citep{jordan2024moddednanogpt,mome}.
However, a single shared table gives every layer the same embedding, leaves no room for layer-wise preferences, and performs poorly in practice.
Early and late layers perform different computations, so a fixed vector suited to one depth need not suit another.
We therefore ask whether one set of n-gram embedding tables can let each layer read what it needs, and how such a design compares empirically with separate tables for each layer.

We propose \method{}, in which one set of n-gram embedding tables serves the whole network and every layer reads it in its own way (\cref{fig:method}b).
In Chorus, n-gram embeddings and token embedding are combined and refined to obtain one refined embedding for the whole network.
Each layer that uses this vector first adapts it to its own needs: it rescales the components of the vector by factors computed from its current hidden state, and then maps the result through its own projection layer.
The rescaling lets a layer emphasize the components of the refined embedding that matter in its current context, and the projection maps them into the form that layer's own computation uses, so the same refined embedding can serve an early layer and a late one differently.

We compare \method{} with existing n-gram methods on a nine-layer model with 20.7M backbone parameters, a GPT-style model from the nanochat codebase \citep{karpathy2025nanochat}, trained on ClimbMix \citep{climbmix} with ten table sizes from 20M to 9.6B (\cref{fig:scaling}).
\method{} reaches a lower validation loss than Engram and FactorEngram at every table size tested.
The margin is large: with 1.2B tables, \method{} predicts text better than either baseline does with 9.6B, so it surpasses their best quality with 87\% smaller tables.
The result carries over to another dataset.
On FineWeb \citep{penedo2024fineweb}, with a nine-layer model of 17.1M backbone parameters, \method{} reaches 1.193 bits per byte against 1.198 for TN-gram, the strongest of four baselines.
It also generalizes to a larger network: in an 18-layer model with 133M backbone parameters, \method{} reaches 1.057 bits per byte against 1.066 for Engram and 1.065 for TN-gram.
We further show that the recently proposed memory-efficient optimizer Ember works well with \method{}, saving 99.2\% of table-optimizer memory compared with AdamW without degrading training performance.

Our contributions are as follows:
\begin{itemize}[leftmargin=1.4em,itemsep=1pt,topsep=2pt]
\item \textbf{A design for sharing n-gram tables across the network.} \method{} combines a per-token table and hashed n-gram tables into one vector used at the input and in the attention values of every other layer, where each layer rescales it from its hidden state and applies its own projection.
\item \textbf{Lower loss at every table size, and the same quality with 87\% smaller tables.} Across ten table sizes from 20M to 9.6B on ClimbMix, \method{} achieves a lower validation loss than Engram and FactorEngram at every size and outperforms their 9.6B models using 1.2B tables.
\item \textbf{The advantage holds across datasets and at larger scale.} \method{} has the lowest loss among seven methods on ClimbMix, outperforms four baselines on FineWeb, and improves over Engram and TN-gram in an 18-layer, 133M-parameter model.
\item \textbf{Memory-efficient training with Ember.} Training \method{}'s tables with Ember saves 99.2\% of their optimizer-state memory compared with AdamW without degrading training performance.
\end{itemize}

\section{\method{}}
\label{sec:method}

\begin{figure}[t]
\centering
\ifdualversehtml
\includegraphics[width=\textwidth]{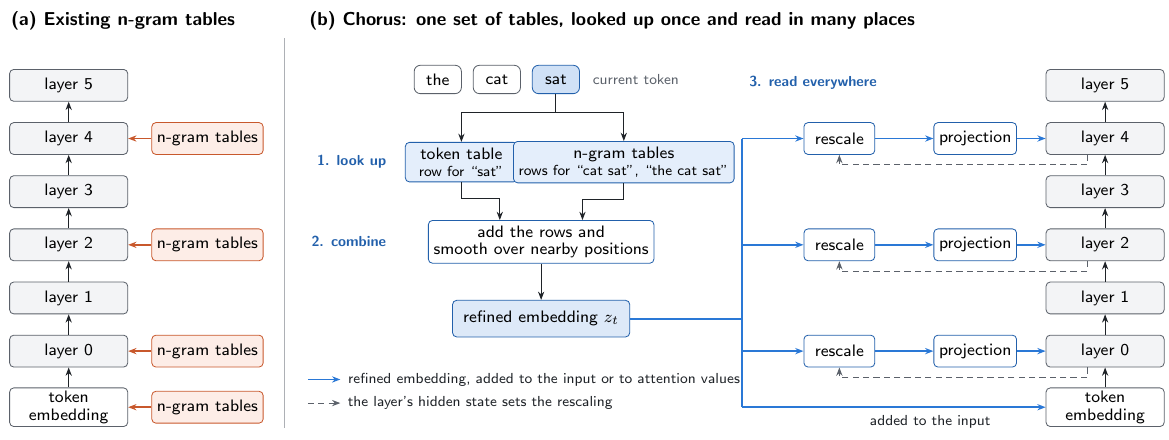}
\else
\resizebox{\linewidth}{!}{%
\begin{tikzpicture}[font=\sffamily\footnotesize, >={Stealth[length=4pt]},
  blk/.style={draw=muted, rounded corners=2pt, minimum width=20mm, minimum height=5.4mm, fill=sitefill, inner sep=1.5pt},
  tab/.style={draw=oursblue!80!black, rounded corners=2pt, fill=tablefill, minimum height=6mm, inner sep=2.5pt, align=center},
  op/.style={draw=oursblue!80!black, rounded corners=2pt, fill=white, minimum height=5.4mm, inner sep=2.5pt, align=center},
  chip/.style={draw=muted, rounded corners=3pt, fill=white, minimum height=4.8mm, minimum width=8mm, inner sep=2pt},
  mem/.style={draw=engramorange!85!black, rounded corners=2pt, fill=engramorange!12, minimum height=5.4mm, inner sep=2.5pt, align=center},
  flow/.style={->, draw=ink, line width=0.5pt}, read/.style={->, draw=oursblue, line width=0.9pt},
  cond/.style={->, draw=muted, line width=0.5pt, densely dashed}]
\node[font=\sffamily\small\bfseries, anchor=west] at (-0.2,6.5) {(a) Existing n-gram tables};
\foreach \i/\y in {0/0.9,1/1.8,2/2.7,3/3.6,4/4.5,5/5.4}{\node[blk] (L\i) at (0.9,\y) {layer \i};}
\foreach \i/\j in {0/1,1/2,2/3,3/4,4/5}{\draw[flow] (L\i) -- (L\j);}
\node[blk, fill=white, align=center] (Lin) at (0.9,-0.05) {token\\[-1pt]embedding}; \draw[flow] (Lin) -- (L0);
\node[mem] (M0) at (3.25,-0.05) {n-gram tables};
\node[mem] (M1) at (3.25,0.9) {n-gram tables};
\node[mem] (M2) at (3.25,2.7) {n-gram tables};
\node[mem] (M3) at (3.25,4.5) {n-gram tables};
\draw[->, draw=engramorange!85!black, line width=0.9pt] (M0) -- (Lin);
\draw[->, draw=engramorange!85!black, line width=0.9pt] (M1) -- (L0);
\draw[->, draw=engramorange!85!black, line width=0.9pt] (M2) -- (L2);
\draw[->, draw=engramorange!85!black, line width=0.9pt] (M3) -- (L4);
\draw[muted!50, line width=0.4pt] (4.55,-0.4) -- (4.55,6.2);
\begin{scope}[xshift=6.95cm]
\node[font=\sffamily\small\bfseries, anchor=west] at (-2.1,6.5) {(b) \method{}: one set of tables, looked up once and read in many places};
\node[chip] (T1) at (0.2,5.5) {the}; \node[chip] (T2) at (1.2,5.5) {cat}; \node[chip, fill=oursblue!22, draw=oursblue!80!black] (T3) at (2.2,5.5) {sat};
\node[font=\sffamily\scriptsize, text=muted, anchor=west] at (2.7,5.5) {current token};
\node[font=\sffamily\scriptsize\bfseries, text=oursblue!80!black, anchor=east] at (-0.55,4.1) {1. look up};
\node[tab, minimum width=19mm] (U) at (0.6,4.1) {token table\\[-1pt]{\scriptsize row for ``sat''}};
\node[tab, minimum width=24mm] (E) at (3.35,4.1) {n-gram tables\\[-1pt]{\scriptsize rows for ``cat sat'', ``the cat sat''}};
\draw[flow] (T3.south) -- ++(0,-0.32) -| (U.north);
\draw[flow] (T3.south) -- ++(0,-0.32) -| (E.north);
\node[font=\sffamily\scriptsize\bfseries, text=oursblue!80!black, anchor=east] at (-0.55,2.75) {2. combine};
\node[op, minimum width=30mm] (R) at (1.95,2.75) {add the rows and\\[-1pt]smooth over nearby positions};
\draw[flow] (U.south) -- ++(0,-0.28) -| ([xshift=-7mm]R.north); \draw[flow] (E.south) -- ++(0,-0.28) -| ([xshift=7mm]R.north);
\node[op, fill=oursblue!16, minimum width=30mm, minimum height=6.2mm] (Z) at (1.95,1.45) {refined embedding $z_t$};
\draw[flow] (R) -- (Z);
\node[font=\sffamily\scriptsize\bfseries, text=oursblue!80!black, anchor=west] at (5.35,5.45) {3. read everywhere};
\foreach \i/\y in {0/0.9,1/1.8,2/2.7,3/3.6,4/4.5,5/5.4}{\node[blk] (B\i) at (11.5,\y) {layer \i};}
\foreach \i/\j in {0/1,1/2,2/3,3/4,4/5}{\draw[flow] (B\i) -- (B\j);}
\node[blk, fill=white, align=center] (Bin) at (11.5,-0.05) {token\\[-1pt]embedding}; \draw[flow] (Bin) -- (B0);
\draw[read, -] (Z.east) -- (5.35,1.45);
\draw[read, -] (5.35,-0.05) -- (5.35,4.5);
\foreach \i/\y in {0/0.9,2/2.7,4/4.5}{
  \node[op, minimum width=12mm] (S\i) at (7.0,\y) {rescale};
  \node[op, minimum width=12mm] (P\i) at (9.3,\y) {projection};
  \draw[read] (5.35,\y) -- (S\i.west);
  \draw[read] (S\i.east) -- (P\i.west);
  \draw[read] (P\i.east) -- (B\i.west);
  \draw[cond] ([xshift=-3mm]B\i.south) -- ++(0,-0.17) -| (S\i.south);
}
\draw[read] (5.35,-0.05) -- node[below, font=\sffamily\scriptsize, text=ink, pos=0.62] {added to the input} (Bin.west);
\draw[read] (-2.0,0.42) -- ++(0.55,0) node[right, font=\sffamily\scriptsize, text=ink] {refined embedding, added to the input or to attention values};
\draw[cond] (-2.0,0.02) -- ++(0.55,0) node[right, font=\sffamily\scriptsize, text=ink] {the layer's hidden state sets the rescaling};
\end{scope}
\end{tikzpicture}}
\fi
\caption{\textbf{Existing n-gram embedding tables and \method{}.} (a) Engram, TN-gram and FactorEngram keep a separate n-gram table for every place that reads one, including the input. (b) \method{} looks up one set of tables once and reuses the result at the input and in every other layer. Each layer rescales the refined embedding using its own hidden state (dashed arrows) and maps it through its own projection.}
\label{fig:method}
\end{figure}

\method{} maintains one set of n-gram tables, shared across the input and every layer, with layer-wise rescaling and projection (\cref{fig:method}b).
Concretely, \method{} performs the following operations.
Let $x_t\in\{0,\dots,V-1\}$ be the token ID at position $t$, $w_t\in\R^d$ its model input embedding in a model of width $d$, and $h^\ell_t$ the hidden state entering layer $\ell$.

\paragraph{Lookup.}
We use hashed n-gram embedding tables as in previous work \citep{huang2025oe,engram,factorengram}.
We use four tables: $E_1,E_2$ are bigram tables, while $E_3,E_4$ are trigram tables.
Each $E_j\in\R^{m_j\times d/4}$ has $m_j=m+2(j-1)$ rows, where $m$ is the base table size.
Besides these n-gram tables, we also keep a token table $U\in\R^{V\times d}$ for encoding the current token.

To encode the n-grams, we first compute the raw indices in base $V+1$ (the additional one is a special token for missing predecessor at sequence start):
\[
k_t^{(2)} = x_t+(V+1)x_{t-1},\qquad
k_t^{(3)} = k_t^{(2)}+(V+1)^2x_{t-2}.
\]
Since the raw indices can be very large, we hash them using a modulo operation so that they fit the table sizes:
\[
a_{t,j} = k_t^{(n_j)}\bmod m_j,
\]
where $n_j$ is the n-gram length for table $E_j$. Finally, we concatenate the four looked-up embeddings and add the token embedding:
\begin{equation}
e_t = U[x_t] \;+\; [\,E_1[a_{t,1}];E_2[a_{t,2}];E_3[a_{t,3}];E_4[a_{t,4}]\,] \;\in\R^d ,
\label{eq:lookup}
\end{equation}
where $[\,\cdot\,;\cdot\,]$ denotes concatenation.

\paragraph{Refinement.}
After computing $e_t$, we refine it similarly to Engram's causal convolution \citep{engram}, but add two gates so that the model can emphasize different channels in different local token contexts. Concretely, we combine the current token embedding and its two predecessors into a local context:
\[ c_t=\tfrac{1}{\sqrt 3}\Big(\bar w_t+\tfrac{1}{\sqrt2}\,\rho_1(\bar w_{t-1})+\tfrac12\,\rho_2(\bar w_{t-2})\Big) . \]
Here $\bar w_t=\norm(\sg(w_t))$, where $w_t$ is the model input embedding (not to be confused with $U$, the token table for n-gram embedding), $\norm$ is RMS normalization \citep{zhang2019rmsnorm}, and $\sg$ denotes stop-gradient. The function $\rho_k:\R^d\to\R^d$ shifts the channels cyclically by $k$ positions.

This mixed embedding $c_t$ is then used to compute the two gates $a_t$ and $b_t$:
\[
[a_t;b_t]=2\sigma(W_c\,c_t)\in(0,2)^{2d}.
\]
Here $W_c\in\R^{2d\times d}$ is a bias-free linear map and $\sigma$ is the logistic function. The gate $a_t$ is used to rescale $e_t$:
\[
u_t=a_t\odot e_t,
\]
where $\odot$ denotes elementwise multiplication. The gate $b_t$ will be used later for input injection.

This $u_t$ is then passed into the convolutional refinement as in Engram to obtain the \emph{refined embedding} $z_t$:
\begin{equation}
z_t \;=\; u_t+\silu\!\big(\conv_4(\norm(u))_t\big).
\label{eq:refine}
\end{equation}
Here $u$ is the sequence of gated vectors. The convolution has width four and unit dilation, with zeros padded on the left at sequence start.
The refinement depends only on token embeddings and table lookups, so $z_t$ is computed once per position and reused at the input and in all selected read layers.

\paragraph{Input injection.}
We project $z_t$ with a learned linear map $W_{\mathrm{in}}\in\R^{d\times d}$, then use the gate $b_t$ to rescale its channels. We add the result to the model input embedding $w_t$ and normalize the sum:
\begin{equation}
h^0_t=\norm\!\big(w_t+b_t\odot W_{\mathrm{in}}z_t\big).
\label{eq:input}
\end{equation}
Here $h^0_t$ is the input to the first model layer.

\paragraph{Layer-wise adaptation.}
Since all layers use the same refined embedding $z_t$, we use layer-wise adaptation to meet the needs of different layers. Specifically, we select the even-numbered layers for insertion, and rescale and project $z_t$ separately at these selected layers before adding it to the attention values.

First, we compute a channel gate $s_t^\ell$ using the layer's input hidden state $h_t^\ell$:
\[
s_t^\ell = 2\sigma\!\big(A_\ell\,\norm(h_t^\ell)\big) \in(0,2)^d,
\]
where $A_\ell\in\R^{d\times d}$ is a learned linear map specific to layer $\ell$.

This gate $s_t^\ell$ is used to rescale $z_t$, after which we project it to obtain the adapted embedding $\tilde z_t^\ell$:
\[
\tilde z_t^\ell=P_\ell(s_t^\ell\odot z_t)\in\R^{d_v},
\]
where $P_\ell\in\R^{d_v\times d}$ is a learned linear map specific to layer $\ell$, and $d_v$ is the total width of the attention values.

Before adding $\tilde z_t^\ell$ to the attention values, we apply an additional gate so that each value head can control how much it receives. This gate is based on nanochat's per-head gate \citep{karpathy2025nanochat}:
\[
g_t^\ell = 2\sigma\big(G_\ell\,[\hat h^\ell_t]_{1:32}\big)\in(0,2)^{H_v}.
\]
Here $\hat h_t^\ell$ is the normalized hidden state used by attention. The learned map $G_\ell\in\R^{H_v\times32}$ uses its first 32 coordinates to produce one gate for each of the $H_v$ value heads. Each gate scales all channels of its value head.

Finally, we add the gated adapted embedding to the attention values $v_t^\ell$ before attention aggregation:
\begin{equation}
v^\ell_t \;\leftarrow\; v^\ell_t+g^\ell_t\odot \tilde z_t^\ell.
\label{eq:read}
\end{equation}
\paragraph{Training the tables.}
To reduce optimizer memory, we propose training the tables with Ember \citep{ember}, a recently proposed memory-efficient optimizer previously applied to token-indexed tables; here, we apply it to hashed n-gram tables. Instead of keeping two moment arrays with the same shape as the table, as in AdamW, Ember stores only one scalar per row and one per column by factoring the second-moment estimate and omitting momentum.

Concretely, Ember computes bias-corrected exponential averages $\hat r_i$ and $\hat c$ of the mean squared gradient in each row $i$ and each column, using decay $\beta=0.95$. Denoting by $E$ any of the tables $U,E_1,\dots,E_4$, and by $G$ its gradient, each row $i$ accessed in the batch is updated as
\begin{equation}
E[i]\;\leftarrow\;E[i]-\eta\,\frac{G[i]}{\sqrt{\hat r_i\;\hat c/\operatorname{mean}(\hat c)}+\epsilon}.
\label{eq:ember}
\end{equation}
Here $\eta$ is the learning rate and $\epsilon$ a numerical stabilizer. Note that the hidden-layer weight matrices still use Muon \citep{jordan2024muon,amsel2025polar}, instead of Ember. For each $R \times C$ table, Ember reduces optimizer-state storage from $2RC$ values with a typical optimizer such as AdamW to $R+C$ values. At a table size of 1.2B, Ember therefore reduces the table-optimizer state from 4.47 GiB to 35.7 MiB, about $128\times$ less (\cref{tab:ember_memory}).

\section{Results}
\label{sec:results}
\label{sec:setup}

We evaluate GPT-style autoregressive Transformers, using a nanochat-based \citep{karpathy2025nanochat} nine-layer, 20.7M-parameter model on ClimbMix \citep{climbmix}, and Parameter Golf models with nine layers (17.1M parameters) or 18 layers (133M parameters) on FineWeb \citep{penedo2024fineweb}. Within each setting, methods share the backbone initialization and data order, and are trained on 3.15B tokens for the nine-layer models or 5.24B tokens for the 18-layer models. We compare with a plain model, OE \citep{huang2025oe}, X-gram \citep{xgram}, Engram \citep{engram}, TN-gram \citep{tngram} and FactorEngram \citep{factorengram}. Engram, TN-gram and FactorEngram use the optimizers suggested in their papers.

To ensure a fair comparison based on the actual number of added parameters, we count the token and n-gram tables together with all other learnable parameters introduced by \method{} (e.g., the per-layer projection weights). Since the embedding tables account for most of these parameters, we call this total number of parameters the \emph{table size}. We report validation bits per byte, with lower values indicating better prediction. Method-comparison tables use three seeds and report means and standard errors; the size curve, component ablations and order sweep use one seed. More experimental details can be found in \cref{app:baselines}.

\begin{table}[t]
\centering
\caption{\textbf{Validation loss across all three settings.} Validation bits per byte (lower is better), mean $\pm$ standard error over three seeds. The plain model is the original backbone trained with no added n-gram tables.}
\label{tab:summary}
\label{tab:primary}
\small
\setlength{\tabcolsep}{7pt}
\renewcommand{\arraystretch}{1.15}
\begin{tabular}{lccc}
\ChorusHeadRow
\raisebox{1.55\baselineskip}{\ChorusHead{Method}} & \ChorusHead{\shortstack{ClimbMix\\9-layer model\\20.7M backbone\\76M tables}} & \ChorusHead{\shortstack{FineWeb\\9-layer model\\17.1M backbone\\26M tables}} & \ChorusHead{\shortstack{FineWeb\\18-layer model\\133M backbone\\60M tables}}\\
\addlinespace[3pt]
Plain model & 0.9969 $\pm$ 0.0007 & 1.2422 $\pm$ 0.0008 & 1.0823 $\pm$ 0.0006 \\
OE & 0.9885 $\pm$ 0.0003 & \textemdash & \textemdash \\
X-gram & 0.9688 $\pm$ 0.0007 & \textemdash & \textemdash \\
Engram & 0.9616 $\pm$ 0.0006 & 1.2051 $\pm$ 0.0002 & 1.0665 $\pm$ 0.0001 \\
TN-gram & 0.9602 $\pm$ 0.0024 & 1.1983 $\pm$ 0.0002 & 1.0648 $\pm$ 0.0006 \\
FactorEngram & 0.9565 $\pm$ 0.0005 & 1.2177 $\pm$ 0.0001 & 1.0768 $\pm$ 0.0005 \\
\textbf{\method{} (ours)} & \textbf{0.9473} $\pm$ \textbf{0.0005} & \textbf{1.1926} $\pm$ \textbf{0.0005} & \textbf{1.0568} $\pm$ \textbf{0.0001} \\
 
\bottomrule
\end{tabular}
\end{table}

\Needspace{14\baselineskip}
\subsection{Evaluation Results}
\label{sec:evaluation}
\label{sec:scaling}
\label{sec:fineweb}
\label{sec:downstream}

To understand how \method{} scales with memory size, we train the nine-layer model on ClimbMix with \method{}, Engram, and FactorEngram at ten table sizes ranging from 20M to 9.6B. We increase table size by increasing row counts while holding each embedding width fixed. The scaling results are shown in \cref{fig:scaling}. \method{} achieves the lowest validation loss at every table size; at 9.6B, it reaches 0.9148 \bpb{}, compared with 0.9258 for FactorEngram and 0.9273 for Engram. Increasing the table size from 4.8B to 9.6B reduces the loss by a further 0.0039 \bpb{}, suggesting that \method{} could benefit from even larger tables. Notably, with only 1.2B tables, \method{} achieves 0.9229 \bpb{}, outperforming all 9.6B baselines with 87\% smaller tables, demonstrating its memory efficiency.

\begin{figure}[t]
\centering
\ifdualversehtml
\includegraphics[width=\textwidth]{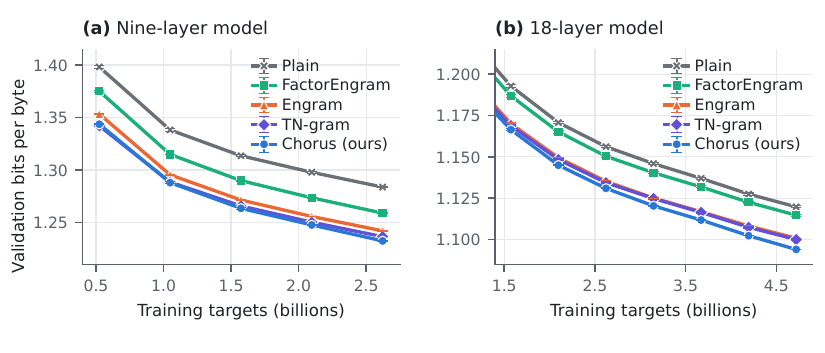}
\else
\includegraphics[width=\linewidth]{figures/fig_fineweb.pdf}
\fi
\caption{\textbf{Training curves at two model sizes.} FineWeb. (a) Nine-layer model (17.1M backbone), 26M tables. (b) 18-layer model (133M backbone), 60M tables. Plain model means no added n-gram tables. Both panels show validation bits per byte on a fixed 2{,}048-sequence subset during training, mean $\pm$ standard error over three seeds. Final full-validation results are in \cref{tab:summary}.}
\label{fig:fineweb}
\end{figure}

Comparisons across settings with multiple seeds show the same advantage (\cref{tab:summary}). Beyond ClimbMix, we test FineWeb using a similar nine-layer model, where \method{} again achieves the lowest validation loss on every seed. We also evaluate the models using a larger 133M-parameter 18-layer backbone with 60M tables; \method{} achieves the lowest validation loss on every seed against the tested baselines. Training curves show that \method{} leads at four of five recorded checkpoints for the nine-layer model, with the sole exception being the first evaluation at update 1000, where TN-gram performs slightly better (\cref{fig:fineweb}). For the 18-layer model, \method{} leads at all nine recorded checkpoints. These results indicate that the performance benefit carries across datasets and model sizes.

\begin{figure}[t]
\centering
\ifdualversehtml
\includegraphics[width=\textwidth]{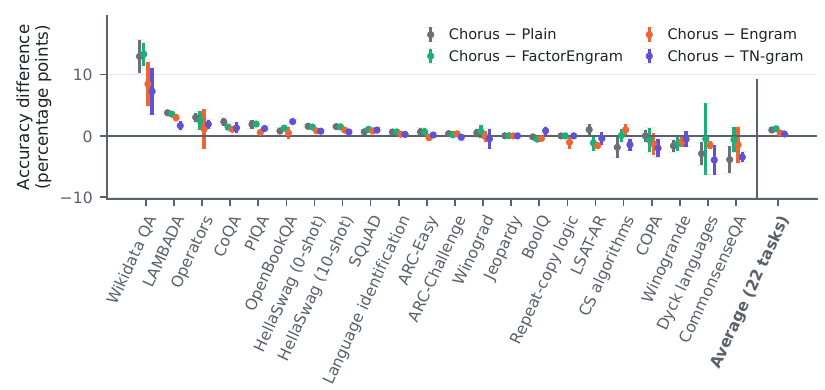}
\else
\includegraphics[width=\linewidth]{figures/fig_downstream.pdf}
\fi
\caption{\textbf{Downstream accuracy.} FineWeb, 18-layer model (133M backbone), 60M tables. Plain model means no added n-gram tables. Accuracy of \method{} minus that of each baseline on 22 tasks and on their average, in percentage points; mean $\pm$ standard error of paired differences over three seeds.}
\label{fig:downstream}
\end{figure}

Beyond validation loss on pretraining data, we evaluate the effect of \method{} on downstream tasks by applying trained 18-layer models to 22 tasks from the DCLM evaluation collection \citep{li2024dclm,karpathy2025nanochat}. Across the full suite of 22 tasks, \method{} achieves the highest mean score of \Scorecodem{}, about 0.0030 ahead of TN-gram, the strongest baseline (\cref{fig:downstream}). These gains are particularly pronounced on Wikidata question answering, where \method{} reaches \Wikidatacodem{}\% accuracy compared to \Wikidataengram{}\% for Engram and \Wikidatatngram{}\% for TN-gram, demonstrating improvements in entity factual recall. We also see improvements on LAMBADA, CoQA, and HellaSwag, which are more closely related to natural-language prediction and comprehension. Chorus has slightly lower accuracy than the baselines on CommonsenseQA and Dyck languages, which are more related to reasoning capability, but neither difference is statistically significant across the three seeds. This pattern suggests that \method{} may help more with factual recall and contextual prediction, likely due to better phrase association. Further downstream evaluation of the smaller model is discussed in \cref{app:downstream}.

\Needspace{8\baselineskip}
\subsection{Optimizer Choice}
\label{sec:optimizer}

\begin{table}[t]
\centering
\caption{\textbf{GPU optimizer memory saved by Ember.} ClimbMix, nine-layer model (20.7M backbone). The state memory is computed using the conventional storage precision for each optimizer.}
\label{tab:ember_memory}
\small
\setlength{\tabcolsep}{5pt}
\begin{tabular}{lrrrr}
\ChorusHeadRow
\ChorusHead{Table size} & \ChorusHead{\shortstack{AdamW state\\BF16 (GiB)}} & \ChorusHead{\shortstack{Ember state\\FP32 (MiB)}} & \ChorusHead{Saved (GiB)} & \ChorusHead{Saved (\%)}\\
\addlinespace[3pt]
76M & 0.27 & 2.11 & 0.27 & 99.25\\
300M & 1.12 & 8.85 & 1.11 & 99.23\\
1.2B & 4.47 & 35.67 & 4.44 & 99.22\\
2.4B & 8.94 & 71.44 & 8.87 & 99.22\\
 
\bottomrule
\end{tabular}
\end{table}

We use Ember instead of conventional AdamW to train n-gram embeddings because it reduces optimizer-state memory during training. Since optimizer states are typically stored on the GPU, this saving allows larger tables to fit on the same GPU. At a table size of 1.2B, Ember requires 35.7 MiB of state compared with 4.47 GiB for AdamW, saving 4.44 GiB (99.2\%; \cref{tab:ember_memory}).

\begin{table}[t]
\centering
\caption{\textbf{The design against the table optimizer, and a simpler shared table.} ClimbMix, nine-layer model (20.7M backbone), 76M tables. Validation bits per byte, mean $\pm$ standard error over three seeds.}
\label{tab:optimizer}
\small
\begin{tabular}{llc}
\ChorusHeadRow
\ChorusHead{Memory} & \ChorusHead{Table update} & \ChorusHead{\bpb{} $\downarrow$ (mean $\pm$ s.e.)}\\
\addlinespace[3pt]
Engram & AdamW & \CtlEngramNativeB\\
Engram & Ember & \CtlEngramEmber\\
FactorEngram & AdamW & \CtlFENative\\
FactorEngram & Ember & \CtlFEEmber\\
Shared bigram table & AdamW & \CtlSharedBigramAdamw\\
Shared bigram table & Ember & \CtlSharedBigram\\
\midrule
\method{} & AdamW & \CtlOursAdamw\\
\method{} (ours) & Ember & \CtlOursEmber\\
 
\bottomrule
\end{tabular}
\end{table}

Beyond memory savings, we examine how optimizer choice affects different n-gram methods. We train \method{}, Engram, FactorEngram, and a simpler shared-bigram design with both AdamW and Ember at a table size of 76M, using three seeds per combination (\cref{tab:optimizer}). Ember slightly improves performance for \method{}, Engram, and the shared-bigram design, but substantially worsens performance for FactorEngram. This suggests that its benefit does not extend to all n-gram methods.

\Needspace{8\baselineskip}
\subsection{Analysis and Ablation Study}
\label{sec:ablation}
\label{sec:controls}

\begin{figure}[t]
\centering
\ifdualversehtml
\includegraphics[width=\textwidth]{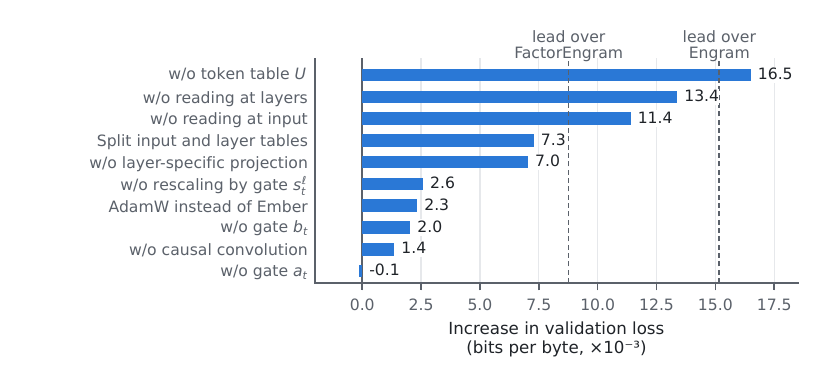}
\else
\includegraphics[width=\linewidth]{figures/fig_controls.pdf}
\fi
\caption{\textbf{Ablation studies of \method{} components.} ClimbMix, nine-layer model (20.7M backbone), 76M tables. Bars show increases in validation loss; bars and reference gaps use one seed. Dashed vertical lines mark \method{}'s gaps to FactorEngram and Engram. Removing the input or layer reads removes the addition of the refined embedding to the token embedding or attention values, respectively. Table size is controlled: in ``w/o token table $U$,'' its parameters are reallocated to the n-gram tables; in ``Split input and layer tables,'' the same total table size is divided equally between the two pools.}
\label{fig:controls}
\end{figure}

We perform ablation analysis on \method{} to understand the importance of each component, as shown in \cref{fig:controls}. Components also used in previous work and baselines are important. For example, the token table $U$ has the largest effect: removing it increases validation loss from 0.9469 to 0.9634 \bpb{}, even when its parameters are reallocated to the n-gram tables. A possible explanation is that dedicated token rows complement the context-dependent phrase features with memory of individual tokens. Reading the refined embedding at both the input and intermediate layers is also important. Removing either read increases loss significantly, showing that the refined embedding provides useful information at both stages of the model.

Splitting the input and layer lookup tables into separate pools also hurts performance. After splitting, each pool gets only half the table size to keep the total table size fixed, which is less efficient than reusing the same table at full size. This suggests that the n-gram tables for the input and intermediate-layer attention contain knowledge that can be shared and reused. Layer-specific projection is also important: sharing a single projection across all layers substantially increases validation loss, indicating that layer-specific adaptation is needed for knowledge reuse.

The other design choices mostly provide modest gains. For example, rescaling by the channel gate $s_t^\ell$ and input gate $b_t$ improves performance moderately, while gate $a_t$ shows no benefit in this comparison. This may reflect overlapping gate functions that make $a_t$ redundant in this setting.

\paragraph{Longest n-gram order.}
To examine how n-gram order affects performance, we vary the longest n-gram order $N$ from 2 to 8; our default is $N=3$. We first fix the total table size, so increasing $N$ divides the same capacity among more n-gram orders. This shows which order performs best under a fixed memory constraint. At 76M table parameters, $N=3$ performs best, matching our default; at 1.2B, $N=4$ performs best, improving BPB over $N=3$ by 0.0009. This suggests that larger tables may benefit from including longer contexts, possibly because they reduce hash collisions. We also consider a growing-size setting: each table keeps the same size as $N$ increases, so the total table size grows from 177.8M at $N=2$ to 1.2B at $N=8$. In this setting, $N=5$ performs best, suggesting that a larger $N$ can still be beneficial when memory is not a constraint.

\begin{figure}[t]
\centering
\ifdualversehtml
\includegraphics[width=\textwidth]{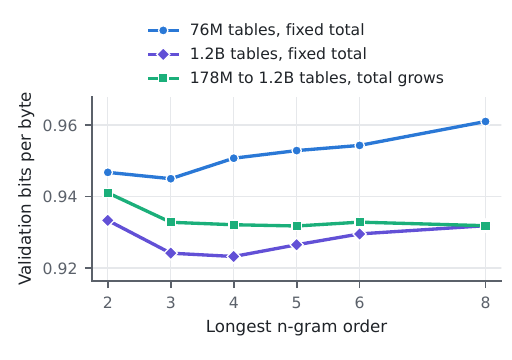}
\else
\includegraphics[width=0.62\linewidth]{figures/fig_nsweep.pdf}
\fi
\caption{\textbf{Effect of the longest n-gram order.} ClimbMix, nine-layer model (20.7M backbone). Validation bits per byte for each longest n-gram order $N$, using all orders from 2 through $N$. Solid lines show fixed total table sizes of 76M and 1.2B. The dashed line shows the growing-size condition: each table keeps the same row count as orders are added, for a total of 177.8M to 1.2B parameters.}
\label{fig:nsweep}
\end{figure}

\paragraph{Phrase frequency.}
We also examine whether the gain is concentrated in rare contexts by grouping validation passages according to how often their three-token contexts occur in training. \method{} has lower loss than Engram and FactorEngram in all four frequency groups at all ten table sizes. Increasing table size from 4.8B to 9.6B improves loss by a similar amount across groups (0.0038--0.0040 \bpb{}), suggesting that the benefit extends to both common and uncommon contexts. Further ablation results and order-sweep details are in \cref{app:controls}.

\section{Related work}
\label{sec:related}

\paragraph{Token and n-gram embedding tables.}
Tensor factors \citep{tngram}, sparse dictionaries \citep{factorengram}, and frequency-based allocation \citep{xgram} improve embedding-table efficiency by changing how entries are represented or allocated within a table. These approaches focus on how entries are represented and allocated within a table. In contrast, Chorus focuses on sharing the same n-gram table throughout the model to save memory and reuse knowledge, complementing such table designs. Recent work also shows that n-gram tables can be stored in low precision to save memory at inference, but not during training \citep{lowbit}.

\paragraph{Sharing embeddings across layers.}
Compact training recipes add token-indexed embeddings to attention values \citep{jordan2024moddednanogpt,karpathy2025nanochat}. MoME uses a context gate to select among multiple vectors per token and also evaluates sharing its token table across memory layers \citep{mome}. Memory Attention constructs values by combining layer-specific token memories with contextual keys \citep{memattn}. Value Residual Learning instead reuses contextual attention values: ResFormer carries first-layer values to deeper layers through residual connections, and SVFormer shares them across all layers \citep{zhou2025valueresidual}. However, these works study different settings from ours, which is n-gram embedding sharing during pre-training.

\paragraph{Optimizers for large tables.}
We use Ember \citep{ember} to train our shared embedding tables. Previous n-gram methods commonly use Adam or AdamW \citep{engram,tngram,factorengram,xgram}. Ember was originally evaluated on input embeddings and language-model output weights, with additional tests on token-indexed value tables. We apply it to hashed n-gram tables and show that it reduces optimizer memory while preserving predictive performance in \method{} and some other n-gram embedding methods.

\section{Conclusion}
\label{sec:conclusion}

In this work, we propose Chorus, an n-gram embedding method that shares tables across the input and intermediate layers while adapting to each layer's needs. Empirical results show improvements over existing baselines across model sizes and training datasets. These results enable further scaling of n-gram embedding tables along the table-size axis by reducing memory use and allowing shared knowledge to be reused across layers.


\bibliographystyle{unsrtnat}
\bibliography{references}

\newpage
\appendix
\raggedbottom

\FloatBarrier
\Needspace{8\baselineskip}
\section{Experiment Details}
\label{app:baselines}

\FloatBarrier
\Needspace{7\baselineskip}
\subsection{Models, data and training}
We use a nanochat-based model on ClimbMix and the Parameter Golf model family on FineWeb, following the setting used by TN-gram \citep{karpathy2025nanochat,climbmix,tngram,penedo2024fineweb}. The main model sizes and training budgets are listed in \cref{tab:settings}. ClimbMix uses an 8,192-token byte-pair vocabulary; FineWeb uses the public tokenized data and 1,024-token SentencePiece vocabulary from Parameter Golf. The models use a 1,024-token context and grouped-query attention with eight query heads and four key/value heads.

\begin{table}[!htbp]
\centering
\caption{\textbf{Model and training settings.} Backbone and table sizes are in millions or billions of parameters. Method comparisons use three seeds; the scaling curve uses one seed at each size.}
\label{tab:settings}
\small
\setlength{\tabcolsep}{4pt}
\begin{tabular}{lccc}
\ChorusHeadRow
\ChorusHead{} & \ChorusHead{ClimbMix} & \ChorusHead{FineWeb, small} & \ChorusHead{FineWeb, large}\\
\addlinespace[3pt]
Layers / width & 9 / 512 & 9 / 512 & 18 / 1,024\\
Backbone parameters & 20.71M & 17.06M & 133.25M\\
Vocabulary size & 8,192 & 1,024 & 1,024\\
Training tokens & 3.15B & 3.15B & 5.24B\\
Table size & 20M--9.6B & 26M & 60M\\
\bottomrule
\end{tabular}
\end{table}

Methods within each setting share the same backbone initialization, data order and training budget. Each update processes 524,288 prediction targets, giving 6,000 updates for the nine-layer models and 10,000 for the 18-layer models. Learning rates remain constant for the first 80\% of training and then decay linearly to zero. The models' original token-indexed value tables are removed in all comparisons. Intermediate training curves use a fixed validation subset; final BPB uses the full validation split. All experiments use RTX 5090 GPUs.

\paragraph{Initialization.}
The token and n-gram tables and the convolution are initialized to zero, so the added memory initially contributes zero to the model. The input projection $W_{\mathrm{in}}$ starts as the identity, and each layer projection $P_\ell$ starts as one quarter of a rectangular identity. The layer gates $A_\ell$ and $G_\ell$ start at zero, giving initial rescaling factors of one; the local-context map $W_c$ uses a small uniform initialization. All these maps and the depthwise convolution are bias-free.

\paragraph{Major optimizer settings.}
We retain the backbone training recipe: Muon with learning rate 0.04 for matrices, Adam with 0.05 for embeddings, and Adam with 0.04 for the remaining backbone parameters. Chorus tables use Ember with learning rate $0.2(d/768)^{-1/2}$, second-moment decay 0.95 and no weight decay. Its projection and gate matrices use Muon with weight decay decaying from 0.2 to zero, and its convolution uses AdamW with learning rate 0.04.

\FloatBarrier
\Needspace{7\baselineskip}
\subsection{Baseline configurations}
We adapt the baselines from their papers or released implementations, retaining their main design and table optimizer. Table row counts or factor ranks are adjusted to the table sizes used in each comparison. Layer numbers below use zero-based indexing.

\paragraph{Engram.}
Engram inserts hashed n-gram memories at selected layers \citep{engram}. On ClimbMix, we use bigrams and trigrams after layers 1 and 4; on FineWeb, we use orders 2--5 after layers 1 and 7. Table sizes are chosen as consecutive primes, lookup uses the model's token vocabulary, and table and other memory parameters use AdamW with learning rates 0.05 and 0.04, respectively.

\paragraph{TN-gram.}
TN-gram uses factorized n-gram memories at selected layers \citep{tngram}. We use orders 2--5 after layers 1 and 7. Factor ranks are 848 on ClimbMix at 76M parameters and 1,409 on nine-layer FineWeb at 26M. The 18-layer result uses a factor rank of 2,252 to match the 60M table size. Factors use AdamW with learning rate 0.05, and key/value projections use Muon.

\paragraph{FactorEngram.}
FactorEngram represents n-gram memories with sparse coefficients over a shared dictionary \citep{factorengram}. We use its unigram, bigram and trigram tables before layers 3 and 4, with four heads per order, width $3d$ and sparsity penalty $10^{-3}$. Table and other memory parameters use AdamW with learning rates 0.001 and 0.01, respectively.

\paragraph{X-gram and OE.}
X-gram's routing tables are built from token frequencies in our training data \citep{xgram}. We use the released configuration with two value-stream views per layer; its larger scaling-curve point additionally uses three hidden-stream views. OE adds four hashed bigram and trigram tables to the input embedding, using the full table budget and Ember for table updates \citep{huang2025oe}.

\FloatBarrier
\Needspace{8\baselineskip}
\section{Additional Results}
\label{app:results}

\FloatBarrier
\Needspace{7\baselineskip}
\subsection{Scaling with table size}
\label{app:scaling}
\Cref{tab:scaling_full} gives the full results behind the scaling curves in \cref{fig:scaling}. In these runs, Chorus uses Ember, while Engram and FactorEngram use AdamW. For table sizes above 2.4B parameters, we use more than one GPU because of the memory limits of our RTX 5090s: two GPUs at 4.8B and four at 9.6B. At 9.6B, each of our four 32-GB GPUs holds one large Chorus n-gram table, whose update requires large temporary buffers. This hardware-constrained placement reduces the peak-memory advantage at the largest size, although Chorus retains its smaller persistent table and optimizer state.

\begin{table}[!htbp]
\centering
\caption{\textbf{Full scaling results.} ClimbMix, nine-layer model (20.7M backbone), one seed per setting. Validation BPB, training-update GPU-hours summed across devices, and peak training memory per GPU.}
\label{tab:scaling_full}
\footnotesize
\setlength{\tabcolsep}{3.5pt}
\begin{tabular}{rlrrrr}
\ChorusHeadRow
\ChorusHead{\shortstack{Table size\\(B parameters)}} & \ChorusHead{Method} & \ChorusHead{GPUs} & \ChorusHead{Validation \bpb{} $\downarrow$} & \ChorusHead{\shortstack{Update\\GPU-hours}} & \ChorusHead{\shortstack{Peak memory\\(GiB/GPU)}}\\
\addlinespace[3pt]
0.020 & \method{} & 1 & \textbf{0.964671} & 1.58 & 8.4 \\
0.020 & FactorEngram & 1 & 0.974093 & 1.87 & 10.0 \\
0.020 & Engram & 1 & 0.973753 & 1.56 & 9.0 \\
\addlinespace[2pt]
0.039 & \method{} & 1 & \textbf{0.955231} & 1.59 & 8.5 \\
0.039 & FactorEngram & 1 & 0.963001 & 1.87 & 10.2 \\
0.039 & Engram & 1 & 0.969229 & 1.56 & 9.1 \\
\addlinespace[2pt]
0.076 & \method{} & 1 & \textbf{0.946917} & 1.60 & 8.7 \\
0.076 & FactorEngram & 1 & 0.955685 & 1.91 & 10.4 \\
0.076 & Engram & 1 & 0.962084 & 1.57 & 9.4 \\
\addlinespace[2pt]
0.153 & \method{} & 1 & \textbf{0.940841} & 1.62 & 9.0 \\
0.153 & FactorEngram & 1 & 0.949911 & 1.89 & 11.0 \\
0.153 & Engram & 1 & 0.952201 & 1.58 & 9.9 \\
\addlinespace[2pt]
0.303 & \method{} & 1 & \textbf{0.935075} & 1.67 & 9.5 \\
0.303 & FactorEngram & 1 & 0.942785 & 1.92 & 12.1 \\
0.303 & Engram & 1 & 0.949177 & 1.60 & 11.1 \\
\addlinespace[2pt]
0.603 & \method{} & 1 & \textbf{0.927558} & 1.76 & 10.7 \\
0.603 & FactorEngram & 1 & 0.939006 & 2.01 & 14.3 \\
0.603 & Engram & 1 & 0.941168 & 1.68 & 13.3 \\
\addlinespace[2pt]
1.203 & \method{} & 1 & \textbf{0.922863} & 1.97 & 12.9 \\
1.203 & FactorEngram & 1 & 0.936389 & 2.22 & 18.9 \\
1.203 & Engram & 1 & 0.937840 & 1.86 & 17.8 \\
\addlinespace[2pt]
2.403 & \method{} & 1 & \textbf{0.920319} & 2.28 & 17.5 \\
2.403 & FactorEngram & 1 & 0.932222 & 2.51 & 27.7 \\
2.403 & Engram & 1 & 0.933491 & 2.19 & 26.7 \\
\addlinespace[2pt]
4.803 & \method{} & 2 & \textbf{0.918736} & 4.85 & 18.6 \\
4.803 & FactorEngram & 2 & 0.929704 & 5.22 & 27.8 \\
4.803 & Engram & 2 & 0.930512 & 4.54 & 26.7 \\
\addlinespace[2pt]
9.603 & \method{} & 4 & \textbf{0.914840} & 10.23 & 27.6 \\
9.603 & FactorEngram & 4 & 0.925836 & 10.60 & 27.8 \\
9.603 & Engram & 4 & 0.927301 & 9.23 & 26.7 \\
 
\bottomrule
\end{tabular}
\end{table}

\begin{figure}[!htbp]
\centering
\includegraphics[width=\linewidth]{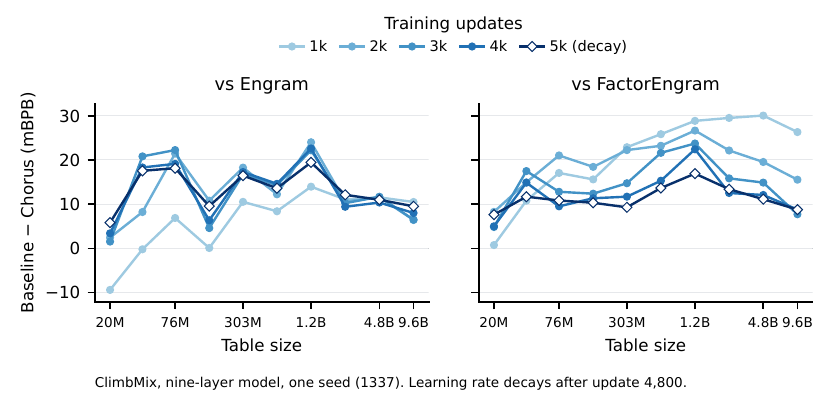}
\caption{\textbf{Validation-loss margins during training.} ClimbMix, nine-layer model (20.7M backbone), one seed per size. Each line is one checkpoint; positive values mean that Chorus has lower validation loss than the baseline. Margins are in thousandths of BPB.}
\label{fig:convergence}
\end{figure}

\FloatBarrier
\Cref{fig:convergence} shows Chorus's lead over Engram and FactorEngram at intermediate checkpoints across the ten table sizes used in \cref{fig:scaling}.

\FloatBarrier
\Needspace{7\baselineskip}
\subsection{Downstream results}
\label{app:downstream}
The 18-layer models are evaluated on the same 91,037 examples across 22 tasks from the DCLM collection, using a nanochat-based evaluator \citep{li2024dclm,karpathy2025nanochat}. The collection covers question answering and factual recall, natural-language prediction and comprehension, commonsense reasoning, and symbolic tasks. Prompts and demonstrations are shared across methods. Standard errors are over three seeds; CommonsenseQA and Dyck comparisons use paired, two-sided $t$-tests at the 5\% level. \Cref{tab:downstream18,tab:downstream18b} give the full task list, example counts and results behind \cref{fig:downstream}.
\newcommand{\DownstreamEighteenColumnCount}{5}
\newcommand{\DownstreamEighteenLayout}{\footnotesize\setlength{\tabcolsep}{3pt}}
\newcommand{\DownstreamEighteenHeader}{\ChorusHeadRow \ChorusHead{Task} & \ChorusHead{Examples} & \ChorusHead{\shortstack{Plain model\\mean $\pm$ s.e.}} & \ChorusHead{\shortstack{FactorEngram\\mean $\pm$ s.e.}} & \ChorusHead{\shortstack{Engram\\mean $\pm$ s.e.}} & \ChorusHead{\shortstack{TN-gram\\mean $\pm$ s.e.}} & \ChorusHead{\shortstack{\method{} (ours)\\mean $\pm$ s.e.}} \\}

\begin{table}[!htbp]
\centering
\caption{\textbf{Eighteen-layer downstream results: tasks 1--11.} FineWeb, 133M backbone, 60M tables. Accuracy (\%), mean $\pm$ standard error over three seeds; bold is best for each task. Plain model means no added n-gram tables.}
\label{tab:downstream18}
\small\DownstreamEighteenLayout
\begin{tabular}{lr*{\DownstreamEighteenColumnCount}{c}}
\DownstreamEighteenHeader
\addlinespace[3pt]
HellaSwag (0-shot) & 10{,}042 & 34.72 $\pm$ 0.12 & 34.84 $\pm$ 0.06 & 35.46 $\pm$ 0.07 & 35.51 $\pm$ 0.19 & \textbf{36.28 $\pm$ 0.17} \\
Jeopardy & 2{,}117 & 0.05 $\pm$ 0.00 & 0.03 $\pm$ 0.02 & \textbf{0.06 $\pm$ 0.02} & \textbf{0.06 $\pm$ 0.02} & \textbf{0.06 $\pm$ 0.02} \\
Wikidata QA & 20{,}321 & 12.88 $\pm$ 0.68 & 12.51 $\pm$ 0.94 & 17.38 $\pm$ 1.20 & 18.57 $\pm$ 1.28 & \textbf{25.84 $\pm$ 2.74} \\
ARC-Easy & 2{,}376 & 35.30 $\pm$ 0.33 & 35.34 $\pm$ 0.14 & \textbf{36.17 $\pm$ 0.04} & 35.79 $\pm$ 0.21 & 35.93 $\pm$ 0.54 \\
ARC-Challenge & 1{,}172 & 21.96 $\pm$ 0.31 & 22.13 $\pm$ 0.42 & 21.99 $\pm$ 0.54 & \textbf{22.55 $\pm$ 0.44} & 22.33 $\pm$ 0.30 \\
COPA & 100 & 59.67 $\pm$ 1.76 & 60.33 $\pm$ 1.86 & 61.00 $\pm$ 1.00 & \textbf{61.67 $\pm$ 0.88} & 59.67 $\pm$ 0.88 \\
CommonsenseQA & 1{,}221 & \textbf{29.29 $\pm$ 1.32} & 26.04 $\pm$ 1.63 & 26.89 $\pm$ 2.36 & 28.88 $\pm$ 1.59 & 25.42 $\pm$ 1.27 \\
PIQA & 1{,}838 & 62.42 $\pm$ 0.34 & 62.37 $\pm$ 0.22 & 63.71 $\pm$ 0.56 & 63.09 $\pm$ 0.13 & \textbf{64.29 $\pm$ 0.46} \\
OpenBookQA & 500 & 24.27 $\pm$ 0.68 & 23.80 $\pm$ 0.42 & 24.60 $\pm$ 0.87 & 22.73 $\pm$ 0.24 & \textbf{25.07 $\pm$ 0.18} \\
LAMBADA & 5{,}153 & 30.53 $\pm$ 0.40 & 30.75 $\pm$ 0.38 & 31.32 $\pm$ 0.04 & 32.64 $\pm$ 0.74 & \textbf{34.30 $\pm$ 0.60} \\
HellaSwag (10-shot) & 10{,}042 & 34.07 $\pm$ 0.07 & 34.10 $\pm$ 0.13 & 34.63 $\pm$ 0.11 & 34.94 $\pm$ 0.07 & \textbf{35.58 $\pm$ 0.11} \\
 
\bottomrule
\end{tabular}
\end{table}

\begin{table}[!htbp]
\centering
\caption{\textbf{Eighteen-layer downstream results: tasks 12--22.} Same models as \cref{tab:downstream18}. Accuracy (\%) and the overall mean score, mean $\pm$ standard error over three seeds; bold is best for each task and for the overall mean.}
\label{tab:downstream18b}
\small\DownstreamEighteenLayout
\begin{tabular}{lr*{\DownstreamEighteenColumnCount}{c}}
\DownstreamEighteenHeader
\addlinespace[3pt]
Winograd & 273 & 56.41 $\pm$ 0.63 & 56.17 $\pm$ 0.74 & 57.02 $\pm$ 0.98 & \textbf{57.39 $\pm$ 1.27} & 56.90 $\pm$ 0.44 \\
Winogrande & 1{,}267 & \textbf{52.46 $\pm$ 0.32} & 52.12 $\pm$ 0.48 & 51.75 $\pm$ 1.08 & 51.35 $\pm$ 0.09 & 50.83 $\pm$ 1.18 \\
Dyck languages & 1{,}000 & 15.33 $\pm$ 1.03 & 12.93 $\pm$ 4.92 & 13.93 $\pm$ 0.79 & \textbf{16.37 $\pm$ 1.27} & 12.43 $\pm$ 1.13 \\
LSAT-AR & 230 & 21.88 $\pm$ 0.29 & 24.06 $\pm$ 0.72 & \textbf{24.49 $\pm$ 0.95} & 23.33 $\pm$ 1.47 & 22.90 $\pm$ 0.63 \\
CS algorithms & 1{,}320 & \textbf{40.98 $\pm$ 0.88} & 39.04 $\pm$ 0.96 & 38.11 $\pm$ 0.20 & 40.56 $\pm$ 1.60 & 39.12 $\pm$ 0.86 \\
Operators & 210 & 14.44 $\pm$ 1.14 & 14.92 $\pm$ 1.24 & 16.35 $\pm$ 2.64 & 15.56 $\pm$ 1.30 & \textbf{17.46 $\pm$ 0.57} \\
Repeat-copy logic & 32 & 0.00 $\pm$ 0.00 & 0.00 $\pm$ 0.00 & \textbf{1.04 $\pm$ 1.04} & 0.00 $\pm$ 0.00 & 0.00 $\pm$ 0.00 \\
SQuAD & 10{,}570 & 1.35 $\pm$ 0.11 & 0.95 $\pm$ 0.10 & 1.21 $\pm$ 0.26 & 1.07 $\pm$ 0.18 & \textbf{2.02 $\pm$ 0.38} \\
CoQA & 7{,}983 & 7.60 $\pm$ 0.80 & 8.46 $\pm$ 0.42 & 8.84 $\pm$ 0.23 & 8.61 $\pm$ 0.72 & \textbf{9.90 $\pm$ 0.20} \\
BoolQ & 3{,}270 & 61.76 $\pm$ 0.27 & \textbf{62.14 $\pm$ 0.06} & 62.03 $\pm$ 0.23 & 60.77 $\pm$ 0.74 & 61.60 $\pm$ 0.23 \\
Language identification & 10{,}000 & 25.49 $\pm$ 0.15 & 25.48 $\pm$ 0.11 & 25.85 $\pm$ 0.05 & 25.88 $\pm$ 0.24 & \textbf{26.12 $\pm$ 0.30} \\
\midrule
\shortstack[l]{Overall mean\\(all 22 tasks)} & & 0.1137 $\pm$ 0.0022 & 0.1119 $\pm$ 0.0005 & 0.1205 $\pm$ 0.0031 & 0.1211 $\pm$ 0.0007 & \textbf{0.1240 $\pm$ 0.0016} \\
 
\bottomrule
\end{tabular}
\end{table}

Separately from the 18-layer DCLM evaluation, we evaluate the nine-layer models on WikiText-2 loss, LAMBADA last-word accuracy and HellaSwag accuracy \citep{merity2017wikitext,paperno2016lambada,zellers2019hellaswag}. These additional results are reported in \cref{tab:external}. WikiText-2 BPB follows each model family's byte-count convention, so comparisons are made within each dataset block.

\begin{table}[!htbp]
\centering
\caption{\textbf{Nine-layer external evaluations.} FineWeb (17.1M backbone, 26M tables) and ClimbMix (20.7M backbone, 76M tables). WikiText-2 BPB and LAMBADA and HellaSwag accuracy, mean $\pm$ standard error over three seeds.}
\label{tab:external}
\small
\begin{tabular}{llccc}
\ChorusHeadRow
\ChorusHead{Setting} & \ChorusHead{Method} & \ChorusHead{\shortstack{WikiText-2 \bpb{} $\downarrow$\\mean $\pm$ s.e.}} & \ChorusHead{\shortstack{LAMBADA (\%) $\uparrow$\\mean $\pm$ s.e.}} & \ChorusHead{\shortstack{HellaSwag (\%) $\uparrow$\\mean $\pm$ s.e.}}\\
\addlinespace[3pt]
FineWeb & Plain model & 1.2813 $\pm$ 0.0019 & 21.50 $\pm$ 0.59 & 29.40 $\pm$ 0.03 \\
 & FactorEngram & 1.2502 $\pm$ 0.0008 & 22.64 $\pm$ 0.26 & 29.65 $\pm$ 0.07 \\
 & Engram & 1.2339 $\pm$ 0.0003 & 22.95 $\pm$ 0.27 & 30.02 $\pm$ 0.18 \\
 & TN-gram & 1.2288 $\pm$ 0.0016 & 24.69 $\pm$ 0.23 & 30.22 $\pm$ 0.11 \\
 & \method{} (ours) & 1.2238 $\pm$ 0.0015 & 24.50 $\pm$ 0.47 & 29.94 $\pm$ 0.15 \\
\midrule
ClimbMix & Plain model & 1.2966 $\pm$ 0.0031 & 19.82 $\pm$ 0.29 & 29.58 $\pm$ 0.01 \\
 & X-gram & 1.2520 $\pm$ 0.0012 & 22.00 $\pm$ 0.30 & 30.71 $\pm$ 0.05 \\
 & FactorEngram & 1.2395 $\pm$ 0.0007 & 21.13 $\pm$ 0.17 & 31.32 $\pm$ 0.15 \\
 & \method{} (ours) & 1.2297 $\pm$ 0.0016 & 23.33 $\pm$ 0.15 & 31.76 $\pm$ 0.06 \\
 
\bottomrule
\end{tabular}
\end{table}

\FloatBarrier
\Needspace{7\baselineskip}
\subsection{Computation and training cost}
\label{app:cost}
\Cref{tab:ops} lists the computation added per token for the FineWeb models. \Cref{tab:cost} gives training time and peak GPU memory for the 18-layer comparisons in \cref{tab:summary}.

\begin{table}[!htbp]
\centering
\caption{\textbf{Computation added per token.} FineWeb models. Counts use multiply-accumulates in matrix products and convolutions; the backbone includes attention over the 1,024-token context and the output projection.}
\label{tab:ops}
\small\ifnum\DownstreamEighteenColumnCount>3\DownstreamEighteenLayout\fi
\begin{tabular}{llcrrr}
\ChorusHeadRow
\ChorusHead{Model} & \ChorusHead{Method} & \ChorusHead{Read sites} & \ChorusHead{Added FLOPs} & \ChorusHead{Backbone FLOPs} & \ChorusHead{Relative (\%)}\\
\addlinespace[3pt]
9 layers, width 512 & Engram & 2 & 2.10M & 21.8M & +9.7\% \\
9 layers, width 512 & TN-gram & 2 & 11.56M & 21.8M & +53.1\% \\
9 layers, width 512 & FactorEngram & 2 & 12.59M & 21.8M & +57.9\% \\
9 layers, width 512 & \method{} & 5 & 2.76M & 21.8M & +12.7\% \\
18 layers, width 1{,}024 & Plain & 0 & 0.00M & 152.0M & +0.0\% \\
18 layers, width 1{,}024 & Engram & 2 & 4.20M & 152.0M & +2.8\% \\
18 layers, width 1{,}024 & TN-gram & 2 & 36.92M & 152.0M & +24.3\% \\
18 layers, width 1{,}024 & FactorEngram & 2 & 50.34M & 152.0M & +33.1\% \\
18 layers, width 1{,}024 & \method{} & 9 & 17.31M & 152.0M & +11.4\% \\
 
\bottomrule
\end{tabular}
\end{table}

\begin{table}[!htbp]
\centering
\caption{\textbf{Training cost.} FineWeb, 18-layer model (133M backbone), 60M tables, on RTX 5090 GPUs. Time spent on training updates and peak training memory, mean $\pm$ standard error over three seeds.}
\label{tab:cost}
\small
\begin{tabular}{lccc}
\ChorusHeadRow
\ChorusHead{Method} & \ChorusHead{\shortstack{Added FLOPs\\(\% of backbone)}} & \ChorusHead{\shortstack{Training time (h)\\mean $\pm$ s.e.}} & \ChorusHead{\shortstack{Peak memory (GB)\\mean $\pm$ s.e.}}\\
\addlinespace[3pt]
Plain model & 0.0 & 17.0 $\pm$ 0.1 & 5.3 $\pm$ 0.0 \\
FactorEngram & 33.1 & 20.5 $\pm$ 0.0 & 6.5 $\pm$ 0.0 \\
Engram & 2.8 & 20.1 $\pm$ 0.4 & 6.0 $\pm$ 0.0 \\
TN-gram & 24.3 & 19.3 $\pm$ 0.0 & 6.5 $\pm$ 0.0 \\
\method{} & 11.4 & 19.3 $\pm$ 0.3 & 6.2 $\pm$ 0.0 \\
 
\bottomrule
\end{tabular}
\end{table}

\FloatBarrier
\Needspace{8\baselineskip}
\section{Additional Analysis}
\label{app:controls}

\FloatBarrier
\Needspace{7\baselineskip}
\subsection{Component ablations}
\Cref{tab:controls_full} gives exact losses and parameter counts for the ablations in \cref{fig:controls}, together with additional configurations that retain the token table but remove all n-gram tables. Additional rows compare the token table with and without layer-wise channel rescaling.

\begin{table}[!htbp]
\centering
\caption{\textbf{Full component ablations.} ClimbMix, nine-layer model, one seed per configuration. Parameter counts are in millions. Rows with a total size of 76.44M reallocate parameters to table rows; other rows also remove the omitted component's parameters. The AdamW row uses learning rate 0.001; the main figure uses the best of three learning rates.}
\label{tab:controls_full}
\small
\begin{tabular}{lrrcc}
\ChorusHeadRow
\ChorusHead{Configuration} & \ChorusHead{\shortstack{Table size\\(M)}} & \ChorusHead{\shortstack{Other\\parameters (M)}} & \ChorusHead{\bpb{} $\downarrow$} & \ChorusHead{\shortstack{$\Delta$ \bpb{}\\vs reference}}\\
\addlinespace[3pt]
\method{} (reference) & 76.44 & 2.75 & 0.946917 & -- \\
Token table only (no n-gram tables) & 6.95 & 2.75 & 0.973654 & +0.026737 \\
w/o token table $U$ and rescaling by gate $s_t^\ell$ & 76.44 & 1.44 & 0.963851 & +0.016934 \\
w/o token table $U$ & 76.44 & 2.75 & 0.963423 & +0.016506 \\
w/o reading at layers & 74.48 & 0.79 & 0.960298 & +0.013381 \\
w/o reading at input & 75.92 & 2.23 & 0.958322 & +0.011405 \\
Split input and layer tables & 76.44 & 2.75 & 0.954209 & +0.007292 \\
w/o layer-specific projection & 75.92 & 2.23 & 0.953961 & +0.007044 \\
AdamW table update (learning rate 0.001) & 76.44 & 2.75 & 0.953312 & +0.006395 \\
w/o rescaling by gate $s_t^\ell$ & 76.44 & 1.44 & 0.949504 & +0.002587 \\
w/o input gate $b_t$ & 76.18 & 2.49 & 0.948960 & +0.002043 \\
w/o convolution & 76.44 & 2.75 & 0.948279 & +0.001362 \\
w/o gate $a_t$ & 76.18 & 2.49 & 0.946769 & -0.000148 \\
 
\bottomrule
\end{tabular}
\end{table}

\FloatBarrier
\Needspace{7\baselineskip}
\subsection{Longest n-gram order}
The order sweep uses two hashed tables for every order from 2 through $N$. A learned projection maps the concatenated lookup to the model width when varying the number of orders. All three curves use this projection at every tested order $N$.

For the fixed-total comparison, rows are divided across the included orders to hold total table size at 76M or 1.2B. For the growing-size comparison, each table retains the row count used at $N=8$, increasing the total from 177.8M parameters at $N=2$ to 1.2B at $N=8$. Each configuration uses one ClimbMix training run.

\FloatBarrier
\Needspace{7\baselineskip}
\subsection{Phrase frequency}
We group validation passages into four equally sized groups according to the frequency of their three-token contexts in the training data, then compute BPB separately in each group. \Cref{tab:phrase_frequency} reports the results for every table size. Chorus outperforms Engram and FactorEngram in every group at all ten table sizes. Increasing table size from 4.8B to 9.6B reduces BPB by 0.0038--0.0040 across groups, showing that the gain extends to both common and uncommon contexts.

\begingroup
\footnotesize
\setlength{\tabcolsep}{7pt}
\begin{longtable}{rcrrr}
\caption{\textbf{Validation loss by phrase frequency.} ClimbMix, nine-layer model (20.7M backbone), one seed per setting. Validation BPB for four equally sized groups of passages, ordered from most frequent three-token contexts (Q1) to least frequent (Q4) in training. Bold is best within each row.}
\label{tab:phrase_frequency}\\
\ChorusHeadRow
\ChorusHead{\shortstack{Table size\\(B parameters)}} & \ChorusHead{Group} & \ChorusHead{Chorus} & \ChorusHead{Engram} & \ChorusHead{FactorEngram}\\
\addlinespace[3pt]
\endfirsthead
\multicolumn{5}{c}{Table \thetable{} (continued)}\\
\ChorusHeadRow
\ChorusHead{\shortstack{Table size\\(B parameters)}} & \ChorusHead{Group} & \ChorusHead{Chorus} & \ChorusHead{Engram} & \ChorusHead{FactorEngram}\\
\addlinespace[3pt]
\endhead
\bottomrule
\endfoot
0.020 & Q1 & \textbf{0.840620} & 0.849840 & 0.850147 \\*
 & Q2 & \textbf{0.941097} & 0.950044 & 0.950421 \\*
 & Q3 & \textbf{0.996503} & 1.005630 & 1.005846 \\*
 & Q4 & \textbf{1.086553} & 1.095585 & 1.096049 \\
\addlinespace[2pt]
0.039 & Q1 & \textbf{0.831665} & 0.845513 & 0.840246 \\*
 & Q2 & \textbf{0.931786} & 0.945722 & 0.939592 \\*
 & Q3 & \textbf{0.986904} & 1.001051 & 0.994468 \\*
 & Q4 & \textbf{1.076635} & 1.090699 & 1.083728 \\
\addlinespace[2pt]
0.076 & Q1 & \textbf{0.823688} & 0.838851 & 0.832810 \\*
 & Q2 & \textbf{0.923603} & 0.938688 & 0.932320 \\*
 & Q3 & \textbf{0.978479} & 0.993654 & 0.987130 \\*
 & Q4 & \textbf{1.067945} & 1.083195 & 1.076514 \\
\addlinespace[2pt]
0.153 & Q1 & \textbf{0.817839} & 0.828915 & 0.827305 \\*
 & Q2 & \textbf{0.917684} & 0.928975 & 0.926783 \\*
 & Q3 & \textbf{0.972389} & 0.983785 & 0.981243 \\*
 & Q4 & \textbf{1.061485} & 1.073173 & 1.070331 \\
\addlinespace[2pt]
0.303 & Q1 & \textbf{0.812327} & 0.825884 & 0.820496 \\*
 & Q2 & \textbf{0.912026} & 0.926216 & 0.919889 \\*
 & Q3 & \textbf{0.966532} & 0.980655 & 0.974068 \\*
 & Q4 & \textbf{1.055433} & 1.069992 & 1.062682 \\
\addlinespace[2pt]
0.603 & Q1 & \textbf{0.804621} & 0.818483 & 0.816667 \\*
 & Q2 & \textbf{0.904651} & 0.918223 & 0.916302 \\*
 & Q3 & \textbf{0.958889} & 0.972400 & 0.970272 \\*
 & Q4 & \textbf{1.048094} & 1.061580 & 1.058770 \\
\addlinespace[2pt]
1.203 & Q1 & \textbf{0.800374} & 0.815389 & 0.814398 \\*
 & Q2 & \textbf{0.900049} & 0.915003 & 0.913761 \\*
 & Q3 & \textbf{0.953927} & 0.969094 & 0.967366 \\*
 & Q4 & \textbf{1.043106} & 1.057874 & 1.056010 \\
\addlinespace[2pt]
2.403 & Q1 & \textbf{0.798173} & 0.811186 & 0.810389 \\*
 & Q2 & \textbf{0.897710} & 0.910851 & 0.909608 \\*
 & Q3 & \textbf{0.951362} & 0.964535 & 0.963242 \\*
 & Q4 & \textbf{1.040016} & 1.053386 & 1.051616 \\
\addlinespace[2pt]
4.803 & Q1 & \textbf{0.796908} & 0.808409 & 0.808094 \\*
 & Q2 & \textbf{0.896170} & 0.907815 & 0.907159 \\*
 & Q3 & \textbf{0.949607} & 0.961608 & 0.960541 \\*
 & Q4 & \textbf{1.038227} & 1.050198 & 1.048982 \\
\addlinespace[2pt]
9.603 & Q1 & \textbf{0.792956} & 0.805439 & 0.804123 \\*
 & Q2 & \textbf{0.892325} & 0.904635 & 0.903478 \\*
 & Q3 & \textbf{0.945822} & 0.958230 & 0.956676 \\*
 & Q4 & \textbf{1.034226} & 1.046873 & 1.045025 \\
 
\end{longtable}
\endgroup

\FloatBarrier
\Needspace{4\baselineskip}
\section{Limitations}
\label{app:limitations}
We tested dense models with at most 133M backbone parameters; larger models and other architectures remain untested. The gains on downstream tasks appear to concentrate on natural-language prediction and knowledge recall, rather than reasoning tasks.

\flushbottom

\ifpreprint\else\input{checklist}\fi
\end{document}